\pdfoutput=1

\documentclass[11pt]{article}

\usepackage[final]{acl}

\usepackage{times}
\usepackage{latexsym}
\usepackage{titlesec}
\titlespacing{\section}{0pt}{12pt}{6pt}

\usepackage[T1]{fontenc}

\usepackage[utf8]{inputenc}

\usepackage{url}
\usepackage{hyperref}

\usepackage{microtype}

\usepackage{tabularx}
\usepackage{inconsolata}
\usepackage{natbib}
\usepackage{graphicx}

\usepackage{notoccite}
\usepackage{booktabs}
\usepackage{cleveref}
\title{Towards Data-Efficient Language Models: A Child-Inspired Approach to Language Learning}

\author{Mohammad Amin Ghanizadeh and Mohammad Javad Dousti\\
Department of Electrical and Computer Engineering, \\College of Engineering, University of Tehran, Tehran, Iran\\
  \texttt{\{ghanizadeh.amin,mjdousti\}@ut.ac.ir}
}

\begin{document}
\maketitle
\begin{abstract}
In this work, we explain our approach employed in the BabyLM Challenge, which uses various methods of training language models (LMs) with significantly less data compared to traditional large language models (LLMs) and are inspired by how human children learn.
While a human child is exposed to far less linguistic input than an LLM, they still achieve remarkable language understanding and generation abilities.
To this end, we develop a model trained on a curated dataset consisting of 10 million words, primarily sourced from child-directed transcripts.
The 2024 BabyLM Challenge initial dataset of 10M words is filtered to 8.5M.
Next, it is supplemented with a randomly selected subset of TVR dataset consisting of 1.5M words of television dialogues.
The latter dataset ensures that similar to children, the model is also exposed to language through media.
Furthermore, we reduce the vocabulary size to 32,000 tokens, aligning it with the limited vocabulary of children in the early stages of language acquisition.
We use curriculum learning and is able to match the baseline on certain benchmarks while surpassing the baseline on others.
Additionally, incorporating common LLM training datasets, such as MADLAD-400, degrades performance.
These findings underscore the importance of dataset selection, vocabulary scaling, and curriculum learning in creating more data-efficient language models that better mimic human learning processes.
\end{abstract}

\section{Introduction}

Language models (LMs) have revolutionized natural language processing, demonstrating remarkable capabilities in understanding and generating human-like text. However, the training of these models typically requires vast amounts of data, often billions of words, which stands in stark contrast to how human children acquire language. The BabyLM Challenge~\cite{babylm_challenge} seeks to bridge this gap by exploring methods to train LMs more efficiently, using significantly less data while still achieving high performance.

Human children develop impressive language skills despite being exposed to far less linguistic input than traditional large language models (LLMs). This observation raises intriguing questions about the efficiency of human language acquisition and the potential for more data-efficient machine-learning approaches. Our research addresses these questions by mimicking the human language acquisition process.

In this work, we present our approach to the BabyLM Challenge, focusing on developing a model that can learn effectively from a dataset more closely aligned with the linguistic experiences of a young child. Our primary contributions are as follows: 

\begin{enumerate}
    \item Dataset curation: We carefully curated a dataset of 10 million words, primarily sourced from child-directed transcripts. This dataset was then refined to 8.5 million words and supplemented with 1.5 million words of television dialogue, acknowledging the role of media exposure in modern language acquisition. 
    \item Vocabulary scaling: To better mimic the limited vocabulary of children in the early stages of language acquisition, we reduced the model's vocabulary size to 32,000 tokens. This constraint forces the model to learn more efficient representations and generalization strategies. Also, this vocab size is similar to the tokenizer used in Llama models~\cite{llama2}.
    \item Model architecture: We adopted the SmolLM model~\cite{SmolLm} which uses a decoder-only Transformer~\cite{few_shot_learners} model with 125 million parameters, trained for 5 epochs. This relatively compact model size allows us to explore the limits of what can be achieved with limited data and computational resources. 
    \item Experimental variations: We conducted several experiments to evaluate the impact of different training strategies: a) We compared model performance with and without the inclusion of television dialogue data. b) We explored the potential benefits of curriculum learning~\cite{curriculum_learning}, developing a method for scoring and sorting data points based on complexity. c) We investigated the impact of incorporating high-quality monolingual datasets, such as MADLAD-400~\cite{madlad}, on model performance. 
\end{enumerate}

The curriculum learning implementation involved developing custom scoring functions to assess sentence complexity based on factors like word count, average word length, unique word ratio, and punctuation usage, similar to~\cite{length_based_cl}. These scores were then used to sort the dataset, allowing for a structured learning progression.

By focusing on dataset selection, vocabulary scaling, and curriculum learning, we present a framework for developing more efficient language models that could have significant implications for both cognitive science and practical NLP applications.

The rest of this paper is as follows. \Cref{sec:methodology} details our methodology. Next, \Cref{sec:experiments} presents our experimental results.  After that, \Cref{sec:discussion} discuss the implications of our findings for future research in data-efficient language model training, and \Cref{sec:conclusion} concludes the paper.

\section{Methodology}
\label{sec:methodology}

Our approach to the BabyLM Challenge involves careful data preparation followed by the implementation of a curriculum learning strategy. This section details our methods for dataset curation and the subsequent application of curriculum learning. 

We started with an initial dataset of approximately 10 million words, primarily sourced from child-directed transcripts. This dataset was chosen to closely mimic the linguistic input that young children typically receive during their language acquisition process. 

To enhance the quality and relevance of our training data, we implemented a rigorous filtering process as explained next:

\textbf{Duplicate removal:} Similar to~\cite{scaling_lm}, we identified and removed exact duplicate sentences from the dataset. This step helps to prevent overfitting to specific phrases and ensures a more diverse linguistic input. 

\textbf{Content refinement:} After duplicate removal, we further refined the dataset based on relevance and quality criteria. For instance, we excluded data points where the ratio of punctuation marks to total words exceeded 0.33 and removed samples with less than 10 characters, resulting in a reduced dataset of approximately 8.5 million words.

Recognizing that modern-age children often acquire language partially through media exposure, we supplemented our refined dataset with television dialogue. We carefully selected approximately 1.5 million words of television dialogue, focusing on content appropriate for and often consumed by young children. 

The TV data from the TVR dataset ~\cite{TVR} was added to our refined 8.5 million word dataset, resulting in a final training corpus of about 10 million words. 
The inclusion of TV dialogue adds diversity to our dataset and better reflects the varied sources of language input in a child's environment.

Following the data preparation phase, we implemented a curriculum learning approach to optimize the training process. This method is designed to present the model with progressively more complex linguistic inputs, mimicking the natural progression of language acquisition observed in human learners. 

We developed a set of scoring functions to assess the complexity of each data point in our dataset. These functions evaluate various linguistic features as briefed next.

\begin{itemize}
\item \textbf{Word count:} A basic measure of sentence length. 

\item \textbf{Average word length:} An indicator of vocabulary complexity. 

\item \textbf{Unique word ratio:} A measure of lexical diversity within a sentence. 

\item \textbf{Punctuation count:} An indirect measure of syntactic complexity.
\end{itemize}

Each data point is passed through these scoring functions, generating a set of individual scores that capture different aspects of linguistic complexity. 

\begin{equation}
score(d) = \sum_{f \in F}^{} w_f f(d),
\end{equation}
\noindent where $d$ is a data point which the score is computed for, $F$ is the set of functions used for scoring, and $w_f$ is the weight of each scoring function, which ranges between 0 and 1. The sum of all weights should be equal to 1.
We conducted experiments with various weight configurations for each function and found that the unique word count function had a greater influence on the final outcome. As a result, we assigned it a weight of $0.4$, while all other functions were assigned a weight of $0.2$.

Once the complexity score is calculated for each data point, we sort the entire dataset in ascending order of these scores. The sorted dataset forms the basis of our curriculum learning approach.

Training begins with the least complex data points (lowest scores).  As training progresses, more complex data points are introduced.  By the end of training, the model has been exposed to the full range of linguistic complexity present in the dataset. Throughout the training process, the model's learning rate decreases. Revisiting simpler examples in later epochs with a lower learning rate helps fine-tune the model's understanding of fundamental concepts while reducing the risk of overfitting.

This gradual exposure to complexity allows the model to build a foundational understanding of simpler linguistic structures before tackling more complex ones, potentially leading to more robust and efficient learning. 

\section{Experiments}
\label{sec:experiments}

\subsection{Experiments' Setup}

To evaluate the effectiveness of our approach in the BabyLM Challenge, we conducted a series of experiments designed to test various aspects of our model\footnote{\url{https://huggingface.co/universitytehran/SmolLM-135M-10M-word}} and training methodology. Our experimental setup was guided by the goal of creating a data-efficient language model that could perform well on benchmark tasks while using significantly less training data than traditional large language models.

We trained a decoder-only transformer model with 125 million parameters. The model was trained for 5 epochs, with the best-performing checkpoint selected based on the model's performance on the  validation dataset. Our vocabulary size was set to 32,000 tokens, aligning with our strategy of mimicking the limited vocabulary of children in the early stages of language acquisition. The employed hyperparameters are summarized in~\Cref{tab:hyperparameters}.

\begin{table}[t]
    \centering
    \begin{tabular}{lr}
      \toprule
      \textbf{Hyperparameter} & \textbf{Value} \\
      \midrule
      Architecture  & SmolLM \\
      Model size   & 125M\\
      Tokenizer vocab size & 32,000\\
       Batch size & 32\\
       Learning rate  & 5e-5\\
       Weight decay& 0.015\\
       Learning rate scheduler & Linear\\
       Number of decoder layers & 30\\
       Number of attention heads & 9\\
       \bottomrule
    \end{tabular}
    \caption{Model and training parameters.}
    \label{tab:hyperparameters}
\end{table}

In the rest of this section, we present the results of these experiments, providing a detailed analysis of our findings and their implications for data-efficient language model training.

\subsection{Results}

In our initial investigation, we explored the impact of utilizing television data as a rich linguistic resource within a constrained data environment. As shown in \Cref{tab:tv_data}, incorporating transcribed text from television shows significantly enhances the model's performance on BLIMP and BLIMP Supplement benchmarks.
We selected 1.5M words from the TVR and MADLAD datasets to replace with those from the original dataset, while keeping the overall dataset size unchanged.
This observation suggests that the diverse language patterns, dialogues, and narratives present in television content provide valuable linguistic information that can be effectively leveraged to improve language model capabilities. 

\begin{table}[t]
    \centering
    \resizebox{\columnwidth}{!}{
    \begin{tabular}{lcc}
         \toprule
         &  \textbf{BLIMP} & \textbf{BLIMP supplement}\\
         \midrule
         Without TV data&  69.8& 57.9\\
         With TV data&  \textbf{72.2} & \textbf{59.1}\\
         MADLAD data &  68.2& 55.0\\
         \bottomrule
    \end{tabular}
    }
    \caption{The impact of adding 1.5M words of training data from TVR and MADLAD datasets on the performance of the model.}
    \label{tab:tv_data}
\end{table}

\begin{table}[ht]
    \centering
    \resizebox{\columnwidth}{!}{
    \begin{tabular}{lcc}
         \toprule
         \textbf{Vocab size} &  \textbf{BLIMP} & \textbf{BLIMP supplement}\\
         \midrule
         30,000 &  71.1 & 57.3\\
         32,000 &  \textbf{72.2} & \textbf{59.1}\\
         50,000 &  69.0 & 54.4\\
         \bottomrule
    \end{tabular}
    }
    \caption{The impact of tokenizer vocabulary size on the performance of the model.}
    \label{tab:tokenizer_comparison}
\end{table}

\begin{table*}[ht]
    \centering
    \resizebox{\textwidth}{!}{
    \begin{tabular}{lccccc}
    \toprule
         &  \textbf{BLiMP} &  \textbf{BLiMP Supplement} &  \textbf{EWoK}  & \textbf{GLUE}  & \textbf{Macro Average}\\
         \midrule
         BabyLlama & 69.8 &  59.5 & \textbf{50.7}  & 63.3  & 60.8 \\
         LTG-BERT&  60.6 & \textbf{60.8}  & 48.9 & 60.3 & 57.7\\
         Ours (w/o curriculum training) & 71.5 & 58.6 & 50.4& 62.8& 60.8\\
         Ours (w/ curriculum training) & \textbf{72.2} & 59.1 & \textbf{50.7} & \textbf{63.9} & \textbf{61.5} \\
         \bottomrule
    \end{tabular}
    }
    \caption{Comparison between our model and baselines on BLiMP~\cite{BLiMP}, BLiMP supplement, GLUE~\cite{GLUE}, and EWoK~\cite{EWOK}.}
    \label{tab:results}
\end{table*}

As shown in \Cref{tab:tokenizer_comparison}, a key finding from our experiments pertains to the optimal vocabulary size for language model training. We discovered that a vocabulary size of approximately 32,000 tokens yields the best-performing models. Interestingly, both smaller and larger vocabulary sizes resulted in diminished performance compared to this optimal range. This finding highlights the importance of carefully considering vocabulary size as a crucial hyperparameter in language model development.

To further validate this observation, we trained our own tokenizer on English language data, specifically targeting a vocabulary size of 32,000 tokens. This custom tokenizer allowed us to tailor the vocabulary to our specific dataset while maintaining the optimal size identified in our experiments. All models trained with tokenizers of various sizes were trained on the same dataset, consisting of 8.5 million samples along with an additional 1.5 million samples from TV data.

The third significant finding from our research demonstrates the efficacy of curriculum learning in boosting model performance. We implemented a curriculum learning approach by assigning scores to each data point in our dataset using the scoring functions discussed earlier in our methodology. By training the model on this scored data, we observed a notable improvement in overall performance.

This curriculum learning strategy enables the model to gradually learn from simpler to more complex examples, potentially leading to more robust and generalizable language understanding. Our results suggest that carefully designed learning curricula can play a crucial role in optimizing the training process and ultimately enhancing the capabilities of language models.

In an effort to explore alternative data sources, we conducted experiments using the MADLAD~\cite{madlad} dataset as a substitute for our initially provided dataset.
For the selection of MADLAD data, we applied the same set of filters used to curate the 8.5 million word dataset. After filtering, we sampled a total of 10 million words from the MADLAD dataset.
Contrary to our expectations, we observed a decrease in performance across both the BLiMP and BLiMP supplement benchmarks. Specifically, the model trained on MADLAD~\cite{madlad} data achieved scores of $68.2$ and $55.0$ on these benchmarks, respectively, which were lower than the scores obtained using our original dataset. 

This unexpected outcome led us to a crucial insight regarding the nature of high-quality data in language modeling. We posit that the definition of \textit{high-quality data} may vary significantly between low-resource and rich-resource language modeling scenarios. In low-resource environments, where data scarcity is a primary constraint, the emphasis may need to be placed on data that is particularly rich in linguistic structures and diverse in its representation of the target language. Conversely, in rich-resource scenarios, the sheer volume of data might compensate for potential variations in quality.

\Cref{tab:results}  compares our model against the baselines. Our model outperforms or matches the baselines across all benchmarks, except for the BLiMP Supplement. Overall, our model's performance exceeds that of the best-scoring baseline. 

\section{Discussion}
\label{sec:discussion}

We hypothesize that data valuation and attribution methods could offer significant advantages over current data selection techniques. While not directly implemented in our study, methods such as Influence functions~\cite{influence_functions} , Representer point~\cite{representer_point}, and dynamic approaches like TracIn~\cite{influence_estimation} and HyDRA~\cite{Hydra}, or RL-based methods for data valuation~\cite{rl_based_data_valuation}, show promise as potential tools for more effective data curation. These techniques, originally designed to quantify the impact of individual data points on model performance, could potentially be adapted to filter large datasets into smaller, higher-quality subsets. Unlike traditional data selection methods such as number of characters~\cite{transfer_learning}, frequency~\cite{multilingual_dataset}, or using a blocklist~\cite{Falcon_LM} that may rely on simplistic criteria, these advanced techniques could provide a more nuanced understanding of data importance. By identifying the most influential or informative samples, they might enable researchers to create more compact yet equally effective training sets. This approach could lead to reduced computational costs, faster training times, and potentially more robust models. Furthermore, in fields where data collection is resource-intensive, such methods might guide more targeted and efficient data gathering strategies. While further research is needed to validate this hypothesis, exploring the application of these methods in data curation could open new avenues for improving the efficiency and effectiveness of machine learning pipelines. 

\section{Limitations}
\label{sec:limitations}
Despite the promising results, this study has several limitations. First, our approach relies on the weights used for scoring data during curriculum learning. With a different set of weights, performance may even decline compared to not using curriculum learning. Furthermore, these weights may vary across different datasets, and finding their near-optimal values could be computationally expensive. 
Second, the appropriate amount of TV data was selected experimentally and may differ for other datasets. Lastly, the effect of training with this procedure on downstream tasks is unclear and may negatively impact model performance in those tasks.
Future research should aim to address these limitations by developing a reliable and robust method for determining score weights, selecting the appropriate portion of TV data, and assessing the influence of this approach on downstream task performance.

\section{Conclusion}
\label{sec:conclusion}

This study, conducted as part of the BabyLM Challenge, has yielded several significant insights into the development of data-efficient language models that more closely mimic human language acquisition. Our approach, focusing on careful dataset curation, vocabulary scaling, and curriculum learning, has demonstrated promising results in training a language model with substantially less data than traditional large language models. 

These results have important implications for both cognitive science and practical NLP applications. By demonstrating that effective language models can be trained on significantly smaller datasets, our work contributes to the ongoing discussion about data efficiency in AI and machine learning. Furthermore, our findings suggest potential avenues for developing more cognitively plausible models of language acquisition, which could inform both AI research and our understanding of human language learning.

\section*{Acknowledgement}
This research was supported in part by the Iran Cognitive Sciences and Technologies Council. We acknowledge the use of OpenAI ChatGPT in the writing and editing process of this manuscript.

\bibliographystyle{acl_natbib.bst}

\bibliography{latex/custom}

\end{document}